 \documentclass[pmlr,twocolumn]{jmlr} 

\usepackage{booktabs}
\usepackage[load-configurations=version-1]{siunitx} 
\usepackage{bm}

\theorembodyfont{\upshape}
\theoremheaderfont{\scshape}
\theorempostheader{:}
\theoremsep{\newline}

\newcommand{\bftab}{\fontseries{b}\selectfont}

\usepackage{caption}

\jmlrworkshop{Machine Learning for Health (ML4H) 2020} 

\title[]{Identifying Decision Points for Safe and Interpretable \titlebreak Reinforcement Learning in Hypotension Treatment}

\author{%
\Name{Kristine Zhang$^*$} \Email{kristine\_zhang@college.harvard.edu}\\
\Name{Yuanheng Wang$^*$} \Email{yuanheng\_wang@g.harvard.edu}\\
\Name{Jianzhun Du$^*$} \Email{jzdu@g.harvard.edu}\\
\Name{Brian Chu} \Email{brianchu@g.harvard.edu}\\
\addr Harvard University
\AND
\Name{Leo Anthony Celi} \Email{leoanthonyceli@yahoo.com}\\
\addr Massachusetts Institute of Technology
\AND
\Name{Ryan Kindle} \Email{kindlr@gmail.com}\\
\addr Massachusetts General Hospital
\AND
\Name{Finale Doshi-Velez} \Email{finale@seas.harvard.edu}\\
\addr Harvard University
}

\begin{document}

\maketitle

\begin{abstract}
Many batch RL health applications first discretize time into fixed intervals.  However, this discretization both loses resolution and forces a policy computation at each (potentially fine) interval.  In this work, we develop a novel framework to compress continuous trajectories into a few, interpretable decision points---places where the batch data support multiple alternatives.  We apply our approach to create recommendations from a cohort of hypotensive patients dataset.  Our reduced state space results in faster planning and allows easy inspection by a clinical expert.
\end{abstract}
\begin{keywords}
batch RL
\end{keywords}

\section{Introduction}
\label{sec:intro}
Many works have investigated reinforcement learning for sequential decision-making in health \citep{yu2019reinforcement}. A standard approach is to take already-collected (batch) data and use that to train a policy.  However, batch RL has many subtleties, and the high stakes of healthcare settings necessitate special attention to safety, interpretability, and robust off-policy evaluation before polices gleaned from batch data are applied to patients \citep{gottesman2019guidelines}.

In this paper, we focus on the issue of time discretization in this batch RL context. To date, most batch RL applications first discretized time into bins and then solved for a policy decision for each bin.  Not only are learnt policies often sensitive to the binning method, but decisions must be solved for every time step, affecting both computation and policy evaluation \citep{tallec2019making}.  


We ask and answer the question: what if we didn't try to solve at every time point?  Indeed, especially with batch data, the \emph{only} times when one might be able to give a recommendation are in states where clinicians treat similar patients differently.  We call these \emph{decision points} (DP).  Compressing trajectories down to high-ambiguity \emph{decision regions}, or areas of the state space with many DPs, has advantages not only for efficient planning but also for interpretability. Summarizing a continuous time and space problem in 20-40 discrete clusters enables clinicians to manually inspect the recommended action for each cluster and thus globally validate a policy. Our contributions are: (1) developing a general pipeline for learning this decision region MDP and (2) demonstrating its interpretability on a hypotension management task.

\section{Related Work}
Many works have considered RL for health, including for hypotension management in the ICU \citep{komorowski18aiclinician, futoma2020popcorn, srinivasan2020hypotension}. Unlike our work, all of these approaches choose some time discretization with a constant time step size and optimize decisions at each of these points.  \citet{gottesman2020interpretable} propose an interpretable RL approach to handling treatment changes or patient measurements with irregular observations, which is similar to our methods, though we leverage the uncertainty of the behavior policy for both state and time compression. \citet{du2020model} take a model-based approach to continuous-time RL with demonstrations in the healthcare environment, whereas it cannot be applied to the batch RL setting.

More broadly, state abstraction (e.g. \citet{li06stateabstraction}) is a popular RL problem. However, these works have been largely in online settings and focused on aggregation based on policy-relevant information like $Q$-values or transitions; they also only compress the state space and the trajectory lengths remain the same.  In the context of batch RL, we instead aim to identify important states for decision-making and filter out regions where the experts have consensus. 

In this way, our methods enable efficient planning and policy validation. \citet{komorowski18aiclinician} also employ clustering to discretize a continuous environment; however, their state clustering methods do not consider practice variation and thus produce large state spaces (750 vs. 32 for us). The issue of time discretization also remains. Other works use deep models for state representations and policies, but these are difficult to interpret, while ours are easy to inspect.


\section{Background and Notation}
A Markov Decision Process (MDP) is a tuple $\langle \mathcal{X}, \mathcal{A}, \mathcal{T}, \mathcal{R}, \gamma \rangle$.  We assume a continuous state space $\mathcal{X} \in \mathbb{R}^d$, discrete action space $\mathcal{A}$, and discount factor $\gamma \in (0, 1]$. $T(x' | x, a)$ and $R(x, a)$ denote the state transition function and reward function, respectively. A policy $\pi:(\mathcal{X}, \mathcal{A}) \to [0, 1]$ defines the probability of an action given a state. The objective is to learn a policy that maximizes the expected return, or sum of discounted rewards $V^\pi = E\left[\sum_t \gamma^t r_t | a_t \sim \pi\right] $.  

In batch reinforcement learning, we cannot interact with the environment during learning, but instead begin with a pre-collected dataset of trajectories $\mathcal{D} = \{(x_i, a_i, x_i', r_i)\}_{i = 1}^N$. These trajectories are sampled by following a (possibly unknown) behavior policy $\pi_b$ and used to learn a new evaluation policy $\pi_e$. MDPs with small discrete state and action spaces can be solved using value or policy iteration \citep{suttonbarto2018}. 

\section{Method}

The innovation of our method is that we do not consider every state to be a potential decision point. Rather, we only consider those points with high behavior policy variability, that is, states whose similar neighbors are administered different treatments. While this idea is simple, its realization requires formalizing many subtleties, which we describe below. 
\begin{figure*}[t]
    \centering
    \includegraphics[width=\textwidth]{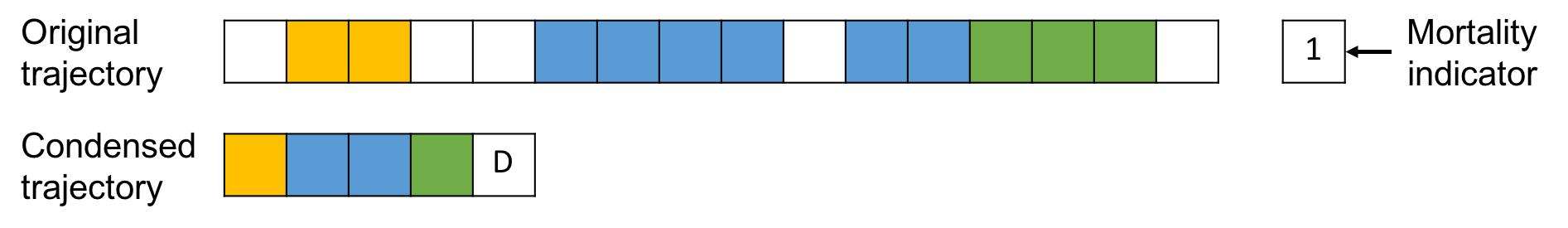}
    \caption[Diagram of MDP compression]{Compression process from a trajectory in the original state space to decision clusters.  White squares are non-DP states, colors denote DPs in the same cluster. The mortality outcome is captured via two more states: A (alive) or D (dead). } 
    \label{fig:mdp_diagram}
\end{figure*}

\paragraph{Decision Point Identification}
We operationalize the notion of states with high clinical variation by first training a kernel-based classifier to \emph{predict} clinician actions.  States with sufficient neighbors (according to the kernel) whose actions still cannot be predicted well are then considered \emph{decision points}.

Specifically, we consider a weighted Gaussian kernel:
\begin{equation}\label{eq:gaussian_kernel}
    k(x, x') = \exp (-\| (x - x') \| _2 ^2)
\end{equation}
where $k(x, x')$ is the estimated similarity between states, and $\bm{w}$ can be interpreted as an importance weighting over state dimensions and $\odot$ represents an element-wise multiplication between vectors. We learn the parameters $\bm{w}$ by backpropagating through the kernel space with optimizing the cross-entropy loss of predicting behavior actions. Details can be found in Appendix \ref{section:kernellearning}. 

The second step is to determine a set of neighbors $\{x'\}$ for each state $x$ in the dataset. We introduce two hyperparameters: the minimum kernel similarity threshold $\delta \in (0, 1]$ to be considered a neighbor, and the minimum number of neighbors $n$ for an action to be considered valid.  That is, if a state $x$ has at least $n$ neighbors within $\delta$ performing action $a$, then action $a$ is considered a valid action for $x$.  If a state has more than one valid action, it is considered a decision point.

\paragraph{Clustering Decision Points}
The process above identifies whether each state in the dataset is a decision point.  Next, we cluster only the decision points to identify discrete \emph{decision regions} that will form our reduced MDP. A top-down hierarchical clustering approach is applied, where the 
Distances between decision points are calculated using the standardized state features (dimensions of $x$). For each intermediate cluster, we examine the difference of mean values for each feature across actions.  If the difference of these means exceed specified thresholds, this implies the region is not homogeneous and we further split clusters (we truly want decision regions where the optimal action is unclear).  
Secondly, we examine if the current clustering is creating loops---defined as trajectories that leave a cluster and coming back to the same cluster within the next three time steps---as these (artificial) loops can make the agent believe that it can ``freeze time'' and avoid any final consequence.  If loop percentage is high, we split that cluster.

\begin{figure*}[t]
\begin{minipage}{0.65\textwidth}
    \centering
    \includegraphics[width=\textwidth]{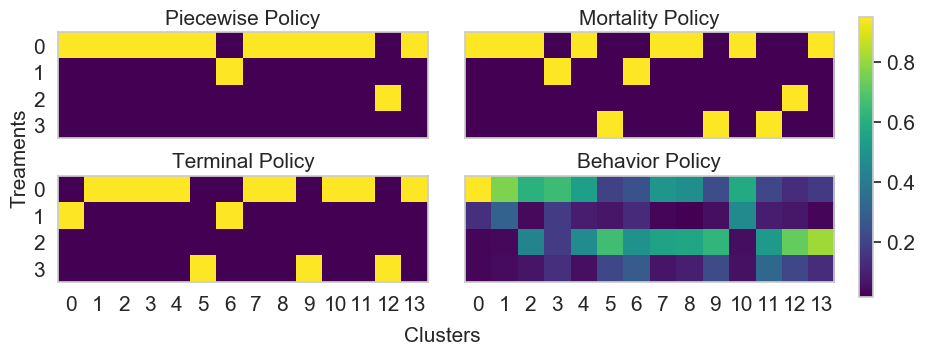}
    \captionof{figure}{Three policies obtained using three different rewards functions, in comparison to the behavior policy. } 
    \label{fig:policy}
\end{minipage}%
\hspace{2mm}
\begin{minipage}{0.3\textwidth}
    \vspace{.75em}
    \setlength{\tabcolsep}{1.5pt}
    \centering
    \begin{tabular}{ccc} 
    \toprule
    Policy & WIS Score & ESS \\
    \midrule
    Behavior & -1.11 & 3,884 \\ 
    Piecewise & \bftab -0.93 & \bftab 1,945 \\
    Mortality & -0.94 & 1,895 \\
    Terminal & -0.94 & 1,921 \\
    \bottomrule
    \end{tabular}
    \vspace{.5em}
    \captionof{table}{OPE results from different policies. ``ESS'' means effective sample size.}
    \label{tab:result}
\end{minipage}
\end{figure*}

\paragraph{Getting a Policy}
Finally, we are ready to create the MDP and solve for a policy. We convert each patient trajectory of the form $\{(x, a, x')\}$ into a new, compressed trajectory $\{(\bar{x}, \bar{a}, \bar{x}')\}$. We use these transitions to learn a discrete, maximum-likelihood MDP in which only valid actions are allowed, and then solve.

The compression process from $\{(x, a, x')\}$ to $\{(\bar{x}, \bar{a}, \bar{x}')\}$ is summarized by Figure \ref{fig:mdp_diagram} and formalized in Appendix \ref{section:mdpcompression}. At a high level, we append a new cluster $\bar{x}_t$ each time a transition is observed from a \textit{different cluster or non-DP state}. Condensing contiguous DPs in the same cluster and ignoring non-DPs is based on the idea that a ``decision'' is an action taken in one decision region that causes movement to another decision region. In the final tuple of a trajectory, we include a transition to the mortality state of the patient to help track their outcome. 

\section{Case Study for Hypotension Management}


\paragraph{Setting} 
Now, we apply our method to recommend treatments to manage hypotension in the ICU.  We select a cohort of 11,739 patients records from the MIMIC-III dataset (See Appendix~\ref{sec:data_processing} for details). Our goal is to (1) recommend a set of treatment policies that are reasonable, and (2) offer a toolkit with visualizations for clinicians to validate and refer to when making a treatment decision for hypotensive patients.  Since this is a proof-of-concept, we only consider four action classes: do nothing, use fluids, use vasopressors, and use both.

To check for sensitivity to reward formulation, we solved with three reward functions: (1) Piecewise: based on the MAP and urine volume for each patient at each timestamp. Lower map and urine values correspond to lower rewards, (2) Mortality probability: inverse of the mortality probability of each cluster-treatment pair in the constructed MDP, and (3) Terminal state: Based on whether patients reach Alive ($R=1$) or Death ($R=0$) at the end of the ICU admission.  We trained the policy using value iteration and use a discount rate of 0.98.

\paragraph{Results}
Running our algorithm on these data results in 33 clusters, where the largest cluster includes 117,730 decision points and the smallest cluster includes 13 decision points. Figure \ref{fig:mean_table} in Appendix \ref{section:dp_clustering} shows the mean value of each feature for clusters containing high mortality rate ($\geq 10\%$) and enough data ($\geq 10$ patients treatment points for at least two types of treatments); this figure was easily inspected by intensivists.  

Figure \ref{fig:policy} shows the optimal policy for each reward function. We also include the behavior policy for comparison, which is the empirical proportion of each action taken in the historical treatment data. The fact that each policy from different rewards functions is slightly different but largely similar to behavior policy demonstrates the robustness of our methods, and the fact it can be so succinctly summarized allowed us to discuss each case with clinicians; in each case, they could posit whether the policy recommendation made sense or could have been due to some confound.  Such discussion would not be possible with a black-box policy.

To quantitatively evaluate the evaluation policy learned using the aforementioned rewards, we use \emph{weighted importance sampling} (WIS) for off policy evaluation (OPE), whose estimate is biased but consistent (asymptotically correct) with much lower variance. We apply standard weight clipping by capping weights at 95 percentile. Table~\ref{tab:result} reports the values of WIS estimates. It shows all three evaluation policies achieve similar rewards, which is higher than the reward of the behavior policy.

\section{Conclusion}
In this work, we present a pipeline for identifying the decision regions in an RL environment and employing them to construct a summary MDP. This summary MDP can then be used for efficient planning and policy validation by human experts. We demonstrate the value of this new approach to interpretability in healthcare settings using the example of hypotension data. Our proposed framework is highly generalizable and can be fine-tuned for other applications in healthcare and beyond. 

\bibliography{jmlr-sample}

\appendix

\section{Data Processing}
\label{sec:data_processing}

\subsection{MIMIC Data Cleaning}

Our hypotension data is obtained from the publicly available MIMIC-III dataset for ICU patients \citep{johnson2016mimic}, with records from 15,653 unique ICU stays. Patients are considered hypotensive if their systolic blood pressure (SBP) drops below 90 mmHg. 

Preprocessing was performed and is described in greater detail by \citet{futoma2020popcorn}. Trajectories were initially discretized into hourly bins starting at $t = 1$ hour into ICU admission and were truncated either at discharge or at $t = 72$ hours (as patients staying beyond 72 hours tend to be abnormal cases). 

The initial state space consists of 111 clinical features including patient measurements, vital signs, and records of past treatment actions. These dimensions contain some redundancy with indicators and quantitative measure for the same variables. 

For actions, we track the amount of either fluids or vasopressors administered to each patient at each time step. For this pipeline, we primarily consider the four general actions corresponding to ``no treatment'', ``fluids only'', ``vasopressors only'', or ``fluids and vasopressors both given''. The learned policy is intended to help physicians decide whether to give each type of treatment at all, as there are existing guidelines for the quantity to give once a choice has been made. Actions are unbalanced in this dataset, with around 80\% of the states receiving no treatment. To analyze treatment impact on outcome, we define a patient's mortality to be true if they died in the hospital within 30 days of ICU admission. 

\subsection{Dataset Split}
Because we lack access to a simulator for MIMIC, the efficacy of our methods must be tested on the batch data. We split the trajectories in the original dataset 75-25 into train and test sets, yielding 555,196 transition tuples from 11,739 trajectories in the train set and 185,671 from 3,914 trajectories in the test set. 

The train set is used for initial data analysis, MDP compression, and planning. The test set is reserved for OPE. 

\subsection{State Feature Selection}

\begin{figure*}[t]
    \centering
    \includegraphics[width=0.75\textwidth]{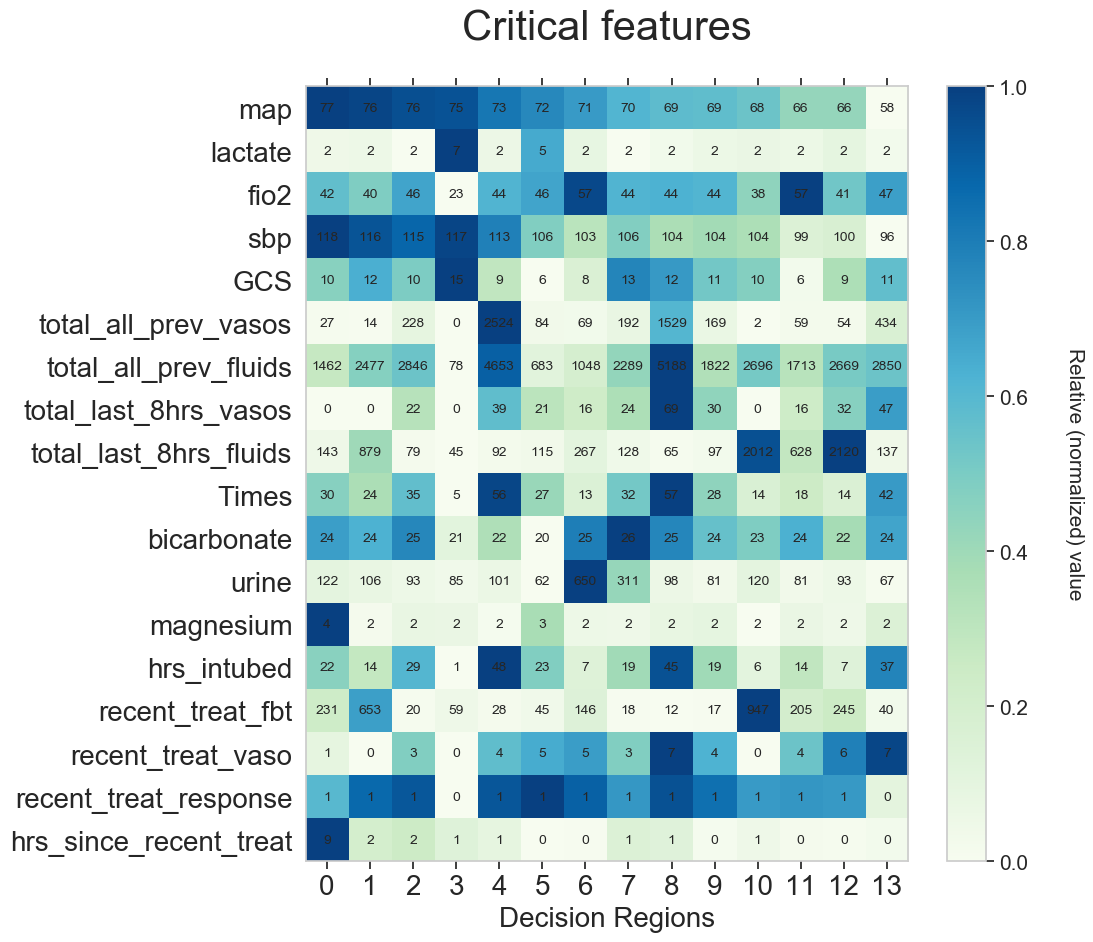}
    \caption[Feature Means for Clusters]{The feature means for points in each decision region. Sorted by descending order of Mean Arterial Pressure (MAP) value.} 
    \label{fig:mean_table}
\end{figure*}

Due to the high dimensionality of the starting MIMIC state space, we preemptively select for features that are most relevant to action prediction. The objective was to increase the efficiency of kernel-learning and avoid collinearity between redundant features. We construct a random forest predictor from states to actions, implemented using a random forest classifier  with 100 estimators, depth 3, and balanced class weights. We employ only the top 20 features ranked by random forest feature importance in the kernel learning algorithm.  Figure \ref{fig:mean_table} lists all continuous features within this top 20 and the features clinicians consider to be important. 

\section{Implementation Details}

\subsection{Kernel Learning}
\label{section:kernellearning}

We are interested in evaluating the Gaussian kernel distance between states, defined as
\begin{equation}
    k(x, x') = \exp (-\| (x - x') \| _2 ^2)
\end{equation}
Naively computing the pairwise distance between $N$ states of dimension $d$ is computationally expensive, given $\mathcal{O}(N^2d)$ complexity and the large MIMIC-III dataset. \citet{rahimi2008random} showed that this process can be accelerated through the use of \emph{Random Fourier Features} (RFF). After applying a randomized feature map $z: \mathbb{R}^d \to \mathbb{R}^D$ from state vectors to a low-dimensional Euclidean inner product space, kernel evaluation can be estimated by the inner product of the randomized features: 
\begin{equation}
    k(x, x') \approx z(x)^\top z(x') 
\end{equation}
We can therefore approximately train kernel machines by projecting to the Fourier space and applying fast linear methods such as regression. We evaluate the efficacy of a set of transformations $\bm{w}$ by performing multi-class logistic regression to predict actions. We define $z(x_i) = RFF(\bm{w} \odot x_i) $ to be the covariates such that $k(x, x') \approx z(x)^\top z(x')$. The targets are $a_i$ with regression coefficients $\bm{V} $. The objective function is the corresponding cross-entropy loss in Equation \ref{eq:cross_entropy}. 
\begin{equation}\label{eq:cross_entropy}
    \begin{gathered}
      \min_{\bm{w}, \bm{V}} -\sum_{i=1}^N \sum_{a \in \mathcal{A}} \mathbb{I}(a_i = a) \hat{p} (a | x_i)  \\ 
      \text{with} \quad \hat{p}(a|x) = \text{softmax}(z(x)^\top \bm{V})_a
    \end{gathered}
\end{equation}
We propose Algorithm \ref{algo:kernel_learning} that uses gradient descent to simultaneously optimize $\bm{w}$ and $\bm{V}$. \citet{rffgradient_nguyen17} showed that the gradient of the cross-entropy loss can be calculated over $\bm{w}$ despite the random feature mapping. We also apply a reparameterization trick, optimizing over $\bm{u} = \log \bm{w}$ to ensure that $\bm{w}$ is positive.

\begin{algorithm2e}[t]
    \SetKwInOut{Input}{Input}\SetKwInOut{Output}{Output}
    \Input{Set of state-action tuples $\{ x_i, a_i \}_{i = 1}^N$; learning rate $\eta$, number of epochs $M$, mini-batch size $B$.}
    \Output{Learned kernel weights $\bm{w}$.} 

    Initialize $\bm{u} \leftarrow \bm{0}^d, \bm{V} \leftarrow \bm{1}^{D \times |\mathcal{A}|}$;\\
    Generate randomized parameters $\omega \in \mathbb{R}^d, b \in \mathbb{R}$ according to the RFF algorithm;\\
    \For{$l \leftarrow 1$ \KwTo $M$}{
        \For{$j \leftarrow 1$ \KwTo $\lfloor N/B \rfloor$} {
            Select mini-batch of states $X_j$ and actions $A_j$;\\
            $\bm{w} \leftarrow \exp(\bm{u})$\\
            Project $z(x) \leftarrow \text{RFF}((\bm{w} \odot x; \omega, b)$ for all $x \in X_j$; \\ 
            $L \leftarrow$ cross-entropy loss when predicting $A_j$ with features $Z_j$ and weights $\bm{V}$; \\
            $\bm{V} \leftarrow \bm{V} - \eta \frac{\partial L}{\partial \bm{V}}$; \\
            $\bm{u} \leftarrow \bm{u} - \eta \frac{\partial L}{\partial \bm{u}}$; \\
        }
    }
    Return $\bm{w} = \exp (\bm{u})$;
    \caption{Learning Weighted Gaussian Kernel}
    \label{algo:kernel_learning}
\end{algorithm2e}

\begin{figure}[htb]
    \centering
    \includegraphics[width=0.3\textwidth]{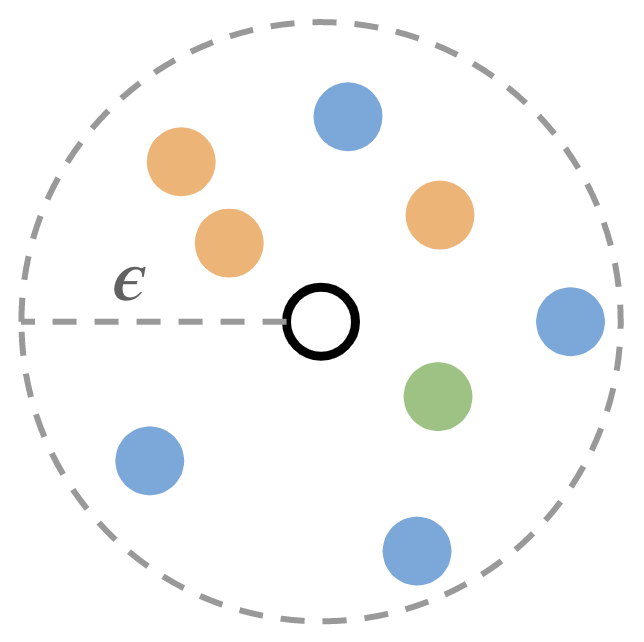}
    \caption{$\epsilon$ conceptually represents distance between points (which inversely approximate similarity $\delta$). Higher similarity threshold$\delta$ translates to a smaller $\epsilon$-ball, which is equivalent to a smaller neighborhood.}
    \label{fig:eps-ball}
\end{figure}

\subsection{DP Hyperparameter Choice}
To identify decision points, we have two main hyperparameters: $\delta$, the minimum similarity from a point $x$ to neighbors, and $n$ the number of neighbors with the same action (Figure~\ref{fig:eps-ball}). Corresponding actions for points present in this $\epsilon$-ball are considered to be allowable actions, and decision points are defined as points that exist in this $\epsilon$-ball where neighbors do not dominantly take a sole action. Since we are more interested to measure the similarity between every point in the kernel space, we instead consider points within the similarity threshold $\delta$. Also, we require a decision point must have a minimum number of neighbors $n$ in this $\delta$-specified region. We perform the grid search over a set of candidate values on a sampled validation dataset. For each pair of $\delta, n$, we evaluate the overall AUC score for a sampled subset of points, whose labels are determined by the empirical distribution of labels present in the region. Conducting such experiment multiple rounds, we finally choose the similarity threshold $\delta=0.95$ and $n=20$ that achieves the highest AUC score. 

\begin{algorithm2e}[htb]
    \SetKwInOut{Input}{Input}\SetKwInOut{Output}{Output}
    \Input{Trajectory $\{ (x_i, a_i) \}_{i=1}^L$ with corresponding cluster labels $\{c_i\}_{i=1}^L$ (with $c_i = 0$ for non-DPs); mortality state $I_m$, action summary function $h$.}
    \Output{Lists of abstracted states and actions $\{\bar{x}_j\}, \{\bar{a}_j\}$. } 
    Initialize $\bar{X} \leftarrow [], \bar{A} \leftarrow [], c_0 \leftarrow 0, A \leftarrow []$  ;\\ 
    \For{$l \leftarrow 1$ \KwTo $L$}{
        \If{$c_l > 0$}{
            $A$.append($a_l$) ;\\
            \emph{// entering new decision region}\\
            \If{$c_l \neq c_0$}{ 
                $\bar{X}$.append($c_l$);
            }
            \emph{// leaving previous decision region} \\
            \If{$l = L$ or $c_{l+1} \neq c_l$} {
                $\bar{A}$.append($h(A)$) ;\\
                $A \leftarrow []$;
            }
        }
    }
    $\bar{X}$.append($I_m$); \\ 
    Return $\bar{X}, \bar{A}$; 
    \caption{MDP Trajectory Compression}
    \label{algo:mdp_compression}
\end{algorithm2e}

\subsection{Decision Region Clustering}
\label{section:dp_clustering}
In order to form decision regions efficiently, we utilize the hierarchical clustering implementations in \texttt{scipy}. We take the output from \texttt{scipy} linkage function call, which is a matrix specifying the order of points being merged at each step. We iterate through the matrix in a top-down manner, and determine if the children of each split meets the specified condition, i.e., the differences of the means of feature values for points in each child are smaller than the predetermined threshold. If not, we continue iterate through the matrix until conditions are satisfied or maximum number of splits are reached.

\subsection{MDP Compression Details}
\label{section:mdpcompression}

Algorithm \ref{algo:mdp_compression} formalizes the MDP compression method. Recall that we convert each patient trajectory $\{(x, a, x')\}$ into a compressed trajectory $\{(\bar{x}, \bar{a}, \bar{x}')\}$, where all $\bar{x}, \bar{x}' \in \mathcal{C}$ are decision clusters and $\bar{a} \in \bar{\mathcal{A}}$ are summarized actions. The summary function $h$ should compress multiple actions in $\mathcal{A}$ that take place while the patient state remains in the same cluster. 

Given a set of compressed trajectories, we estimate a behavior policy $\bar{\pi}_b$ and transition function $\bar{T}(\bar{x}'|\bar{x}, \bar{a})$ from the new dataset, as formalized in Equation \ref{eq:compress_policy} and \ref{eq:compress_transition}. Following standard state abstraction protocol \citep{li06stateabstraction}, the behavior policy is estimated as the proportion of times an action was taken from a given cluster, and the transition function is estimated as the proportion of times a given cluster-action pair led to another cluster. 
\begin{equation}\label{eq:compress_policy}
    \bar{\pi}_b(\bar{a}^* | \bar{x}^*) = \frac{\sum_{i} \mathbb{I}(\bar{x}_i = \bar{x}^*,\ \bar{a}_i =\bar{a}^*) }{\sum_{i} \mathbb{I}(\bar{x}_i = \bar{x}^*)} 
\end{equation}
\begin{equation}\label{eq:compress_transition}
\bar{T}(\bar{x}'^*  | \bar{x}^*, \bar{a}^* )  = \frac{\sum_{i} \mathbb{I} (\bar{x}_i = \bar{x}^*,\ \bar{a}_i = \bar{a}^* ,\ \bar{x}' = \bar{x}'^*) }{\sum_{i} \mathbb{I} (\bar{x}_i = \bar{x}^*,\ \bar{a}_i = \bar{a}^*)} 
\end{equation}
This process can be thought of as simultaneous state and temporal abstraction. We perform state abstraction to map decision points into high-uncertainty clusters, then temporal abstraction to condense all actions that do not cause transitions out of decision regions. In this way, we reduce both the size of the state space and the length of trajectories.

\end{document}